\documentclass[runningheads]{llncs}
\usepackage[T1]{fontenc}
\usepackage[T1]{fontenc}
\usepackage{amsmath,amssymb,amsfonts}
\usepackage{graphicx}
\usepackage{xcolor}
\usepackage{paralist}
\usepackage{caption}
\usepackage{subcaption}
\usepackage{booktabs}
\usepackage{hyperref}
\usepackage{multirow}
\usepackage[hyphenbreaks]{breakurl}

\begin{document}
\title{Architecturing Binarized Neural Networks for Traffic Sign Recognition\thanks{This work was supported by a grant of the Romanian National Authority for Scientific Research and Innovation, CNCS/CCCDI - UEFISCDI, project number PN-III-P1-1.1-TE-2021-0676, within PNCDI III.}}
\titlerunning{Architecturing BNNs for TSR}
\author{Andreea Postovan \and M\u{a}d\u{a}lina Era\c{s}cu}
\authorrunning{Postovan and Era\c{s}cu}
\institute{Faculty of Mathematics and Informatics, West University of Timisoara \\ 
           4 blvd. V. Parvan, 300223, Romania\\
\email{\{andreea.postovan99, madalina.erascu\}@e-uvt.ro}}
\maketitle              
\begin{abstract}


Traffic signs support road safety and managing the flow of traffic, hence are an integral part of any vision system for autonomous driving. While the use of deep learning is well-known in traffic signs classification due to the high accuracy results obtained using convolutional neural networks (CNNs) (state of the art is 99.46\%), little is known about binarized neural networks (BNNs). Compared to CNNs, BNNs reduce the model size and simplify convolution operations and have shown promising results in computationally limited and energy-constrained devices which appear in the context of autonomous driving. 

This work presents a bottom-up approach for architecturing BNNs by studying characteristics of the constituent layers. These constituent layers (binarized convolutional layers, max pooling, batch normalization, fully connected layers) are studied in various combinations and with different values of kernel size, number of filters and of neurons by using the German Traffic Sign Recognition Benchmark (GTSRB) for training. As a result, we propose BNNs architectures which achieve more than $90\%$ for GTSRB (the maximum is $96.45\%$) and an average greater than $80\%$ (the maximum is $88.99\%$) considering also the Belgian and Chinese datasets for testing. The number of parameters of these architectures varies from 100k to less than 2M. The accompanying material of this paper is publicly available at \url{https://github.com/apostovan21/BinarizedNeuralNetwork}.
\end{abstract}
\section{Introduction}\label{sec: Introduction}
Traffic signs are important both in city and highway driving for supporting road safety and managing the flow of traffic. Therefore, \emph{traffic sign classification (recognition)} is an integral part of any vision system for autonomous driving. It consists of: 
\begin{inparaenum}[\itshape a)\upshape]
	\item isolating the traffic sign in a bounding box, and 
	\item classifying the sign into a specific traffic class. 
\end{inparaenum}
This work focuses on the second task. 

Building a traffic sign classifier is challenging as it needs to cope with complex real-world traffic scenes. A well-know problem of the classifiers is the lack of \emph{robustness} to \emph{adversarial examples}~\cite{szegedy2013intriguing} and to occlusions~\cite{zhang2020lightweight}. \emph{Adversarial examples} are traffic signs taken as input which produce erroneous outputs and, together with \emph{occlusions}, they naturally occur because the traffic scenes are unique in terms of weather conditions, lighting, aging. 

One way to alleviate the lack of robustness is to formally verify that the trained classifier is robust to adversarial and occluded examples. 
For constructing the trained model, binary neural networks (BNNs) have shown promising results~\cite{hubara2016binarized} even in computationally limited and energy-constrained devices which appear in the context of autonomous driving. BNNs are neural networks (NNs) with weights and/or activations binarized and constrained to~$\pm 1$. Compared to NNs, they reduce the model size and simplify convolution operations utilized in image recognition task. 

Our long term goal, which also motivated this work, is to give formal guarantees of properties (e.g. robustness) which are true for a trained classifier. The formal \emph{verification problem} is formulated as follows: given a trained model and a property to be verified for the model, does the property hold for that model? To do so, the model and the property are translated into a constrained satisfaction problem and use, in principle, existing tools to solve the problem \cite{10.1007/978-3-540-78800-3_24}. However, the problem is \textsc{NP}-complete~\cite{katz2017reluplex}, so experimentally beyond the reach of general-purpose tool.

This work makes an attempt to arrive at BNN architectures specifically for traffic signs recognition by making an extensive study of variation in accuracy, model size and number of parameters of the produced architectures. In particular, we are interested in BNNs architectures with high accuracy and small model size in order to be suitable in computationally limited and energy-constrained devices but, at the same time, reduced number of parameters in order to make the verification task easier. A bottom-up approach is adopted to design the architectures by studying characteristics of the constituent layers of internal blocks. These constituent layers are studied in various combinations and with different values of kernel size, number of filters and of neurons by using the German Traffic Sign Recognition Benchmark (GTSRB) for training. For testing, similar images from GTSRB, as well as from Belgian and Chinese datasets were used.

As a result of this study, we propose the network architectures (see Section~\ref{sec:expResults}) which achieve more than $90\%$ for GTSRB ~\cite{houben2013detection} and an average greater than $80\%$ considering also the Belgian~\cite{BelgianTrafficSignDatabase} and Chinese~\cite{ChineseTrafficSignDatabase} datasets, and for which the number of parameters varies from 100k to 2M.
\section{Related Work}\label{sec:Related Work}
\begin{figure}[h]
\centering
\includegraphics[width=0.5\textwidth]{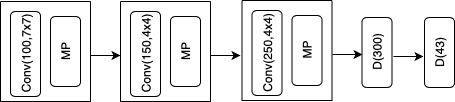}
\caption{\mbox{Architecture for recognizing traffic signs ~\cite{ciregan2012multi}. Image sz: 48$\times$48 \ (px $\times$ px)}}
\label{fig:ArchitectureCiresan}
\end{figure}

\noindent \emph{Traffic Sign Recognition using CNNs.} Traffic sign recognition (TSR) consists in predicting a label for the input based on a series of features learned by the trained classifier. CNNs were used in traffic sign classification since long time ago \cite{sermanet2011traffic,ciregan2012multi}. These works used GTSRB~\cite{houben2013detection} which is maintained and used on a large scale also nowadays. Paper~\cite{ciregan2012multi} obtained
an accuracy of 99.46\% on the test images which is better than the human performance of 98.84\%, while \cite{sermanet2011traffic} with 98.31\% was very close. These accuracies were obtained either modifying traditional models for image recognition (e.g. ResNet in case of \cite{sermanet2011traffic}) or coming up with new ones (e.g. multi-column
deep neural network composed of 25 CNNs in case of \cite{ciregan2012multi}). The architecture from \cite{ciregan2012multi} (see Figure~\ref{fig:ArchitectureCiresan}) contains a number of parameters much higher than those of the models trained by us and it is not amenable for verification although the convolutional layers would be quantized. The work of \cite{ciregan2012multi} is still state of the art for TSR using CNNs.

\noindent \emph{Binarized Neural Networks Architectures.} Quantized neural networks (QNNs) are neural networks that represent their weights and activations using low-bit integer variables. 
There are two main strategies for training QNNs: \emph{post-training quantization} and \emph{quantization-aware training}~\cite{krishnamoorthi2018quantizing} (QAT). The drawback of the post-training quantization is that it typically results in a drop in the accuracy of the network with a magnitude that depends on the specific dataset and network
architecture. In our work, we use the second approach which is implemented in Larq library~\cite{geiger2020larq}. In QAT, the imprecision of the low-bit fixed-point arithmetic is modeled already during the training process, i.e., the network can adapt to a quantized computation during training. The challenge for QNNs is that they can not be trained directly with stochastic gradient descent (SGD) like classical NNs. This was solved by using the straight-through gradient estimator (STE) approach~\cite{hubara2017quantized} which, in the forward pass of a training step, applies rounding operations to computations involved in the QNN (i.e. weights, biases, and arithmetic operations) and in the backward pass, the rounding operations are removed such that the error can backpropagate through the network. 

BinaryConnect~\cite{courbariaux2015binaryconnect} is one of the first works which uses 1-bit quantization of weights during forward and backward propagation, but not during parameter update to maintain accurate gradient calculation during SGD. As an observation, the models used in conjuction with BinaryConnect use only linear layers which is suficient for MNIST~\cite{lecun1998mnist} dataset, but convolutional layers for CIFAR-10~\cite{krizhevsky2009learning} and SVHN~\cite{netzer2011reading}. Paper \cite{hubara2016binarized} binarizes the activations as well. Similarly, for MNIST dataset they use linear layers, while for CIFAR-10, SVHN and ImageNet~\cite{deng2009imagenet} they use variants of ConvNet, inspired by VGG~\cite{simonyan2014very}, with the binarization of the activations.

In \textsc{XNOR}-Net \cite{rastegari2016xnor}, both the weights and the inputs to the convolutional and fully connected layers are approximated with binary values which allows an efficient way of implementing convolutional operations. The paper uses ImageNet dataset in experiments. We use \textsc{XNOR}-Net architectures in our work but for a new dataset, namely traffic signs.

Research on BNNs for traffic sign detection and recognition is scarce. Paper~\cite{chen2021investigating} uses the binarization of RetinaNet~\cite{lin2017focal} and ITA~\cite{chen2019investigating} for traffic sign detection, in the first phase, and then recognition. Differently, we focus only on recognition, hence the architectures used have different underlying principles.

\noindent \emph{Verification of Neural Networks.} Properties of neural networks are subject to verification. In the latest verification competition there are various benchmarks subject to verification \cite{BenchmarksVNN-COMP-22}, however, there is none involving traffic signs. This is because a model with reasonable accuracy for classification task must contain convolutional layers which leads to an increase of number of parameters. To the best of our knowledge there is only one paper which deals with traffic signs datasets \cite{guo2023occrob} that is GTSRB. However, they considered only subsets of the dataset and their trained models consist of only fully connected layers with ReLU activation functions ranging from 70 to 1300. They do not mention the accuracy of their trained models. BNNs \cite{narodytska2018formal,amir2021smt} are also subject to verification but we did not find works involving traffic signs datasets.
\section{Binarized Neural Networks}\label{sec:BNNs}
A BNN \cite{hubara2016binarized} is a feedforward network where weights and activations are mainly binary. \cite{narodytska2018formal} describes BNNs as sequential composition of blocks, each block consisting of linear and non-linear transformations. One could distinguish between \emph{internal} and \emph{output blocks}. 

There are typically several \emph{internal blocks}. The layers of the blocks are chosen in such a way that the resulting architecture fulfills the requirements of accuracy, model size, number of parameters, for example. Typical layers in an internal block are:
\begin{inparaenum}[\itshape 1)\upshape]
\item linear transformation (\textsc{LIN}), 
\item binarization (\textsc{BIN}),
\item max pooling (\textsc{MP}), 
\item batch normalization (\textsc{BN}).
\end{inparaenum}
A linear transformation of the input vector can be based on a fully connected layer or a convolutional layer. In our case is a convolution layer since our experiments have shown that a fully connected layer can not synthesize well the features of traffic signs, therefore, the accuracy is low. The linear transformation is followed either by a binarization or a max pooling operation. Max pooling helps in reducing the number of parameters. One can swap binarization with max pooling, the result would be the same. We use this sequence as Larq~\cite{geiger2020larq}, the library we used in our experiments, implements convolution and binarization in the same function. Finally, scaling is performed with a batch normalization operation~\cite{ioffe2015batch}. 

There is \emph{one output block} which produces the predictions for a given image. It consists of a dense layer that maps its input to a vector of integers, one for each output label class. It is followed by function which outputs the index of the largest entry in this vector as the predicted label.

We make the observation that, if the MP and BN layers are omitted, then the input and output of the internal blocks are binary, in which case, also the input to the output block. The input of the first block is never binarized as it drops down drastically the accuracy.
\section{Datasets and Experimental Setting}\label{sec:DatasetsAndExperimentalSetting}
We use GTSRB \cite{GTSRB} for training and testing purposes of various architectures of BNNs. These architectures were also tested with the Belgian data set \cite{BelgianTrafficSignDatabase} and the Chinese \cite{ChineseTrafficSignDatabase}.

GTSRB is a multi-class, single-image dataset. The dataset consists of images of German road signs in 43 classes, ranging in size from 25 $\times$ 25 to 243 $\times$ 225, and not all of them are square. Each class comprises 210 to 2250 images including prohibitory signs, danger signs, and mandatory signs. The training set contains 39209 images; the remaining 12630 images are selected as the testing set. For training and validation the ratio 80:20 was applied to the images in the train dataset. GTSRB is a challenging dataset even for humans, due to perspective change, shade, color degradation, lighting conditions, just to name a few.

The \emph{Belgium Traffic Signs} test dataset contains 7095 images of 62 classes out of which only 23 match the ones from GTSRB. From that dataset we have used only the images from training folder which are 4533 in total.
The \emph{Chinese Traffic Signs} test dataset contains 5998 traffic sign images for testing of 58 classes out of which only 15 match the ones from GTSRB. For our experiments, we performed the following pre-processing steps on the Belgium and Chinese datasets, otherwise the accuracy of the trained model would be very low: 
\begin{inparaenum}[\itshape 1)\upshape]
	\item we relabeled the classes from the Belgium, respectively Chinese, datasets such that their common classes with GTSRB have the same label, and
	\item we eliminated the classes not appearing in GTSRB.
\end{inparaenum}

In the end, for testing, we have used 1818 images from the Belgium dataset and 1590 from the Chinese dataset.

\smallskip

For this study, the following points are taken into consideration.
\begin{enumerate}
\item Training of network is done on Intel Iris Plus Graphics 650 GPU using Keras v2.10.0, Tensorflow v2.10.0 and Larq v0.12.2. 
\item From the open-source Python library Larq \cite{geiger2020larq}, we used the function \\ \noindent \texttt{QuantConv2D} in order to binarize the convolutional layers except the first. Subsequently, we denote it by QConv. The \texttt{bias} is set to \texttt{False} as we observed that does not influence negatively the accuracy but it reduces the number of parameters. 
\item Input shape is fixed either to $30 \times 30$, $48 \times 48$, or $64 \times 64$ (px $\times$ px). Due to lack of space, most of the experimental results included are for $30 \times 30$, however all the results are available at \url{https://github.com/apostovan21/BinarizedNeuralNetwork}.
\item Unless otherwise stated, the number of epochs used in training is $30$.
\item Throughout the paper, for max pooling, the kernel is fixed to non-overlapping $2 \times 2$ dimension.
\item Accuracy is measured with variation in the number of layers, kernel size, the number of filters and of neurons of the internal dense layer. Various combination of the following values considered are:
\begin{inparaenum}[\itshape (a)\upshape]
\item Number of blocks: $2, 3, 4$;
\item Kernel size: $2, 3, 5$;
\item Number of filters: $16, 32, 64, 128, 256$;
\item Number of neurons of the internal dense layer: $0, 64, 128, 256, 512, 1024$.
\end{inparaenum}
\item ADAM is chosen as the default optimizer for this study. For initial training of deep learning networks, ADAM is the best overall choice \cite{ruder2016overview}.
\end{enumerate}
Following section discusses the systematic progress of the study.
\section{Proposed Methodology}\label{sec:proposedMEthodology}
We recall that the goal of our work is to obtain a set of architectures for BNNs with high accuracy but at the same time with small number of parameters for the scalability of the formal verification. At this aim, we proceed in two steps. First, we propose two simple two internal blocks XNOR architectures\footnote{An \textsc{XNOR} architecture \cite{rastegari2016xnor} is a deep neural network where both the weights and the inputs to the convolutional and fully connected layers are approximated with binary values.} (Section~\ref{sec:XNOR-Architectures}). We train them on a set of images from GTSRB dataset and test them on similar images from the same dataset. We learned that MP reduces drastically the accuracy while the composition of a convolutional and binary layers (QConv) learns well the features of traffic signs images. In Section~\ref{sec:TwoInternalBlocks}, we restore the accuracy lost by adding a BN layer after the MP one. At the same time, we try to increase the accuracy of the architecture composed by blocks of the QConv layer only by adding a BN layer after it.

Second, based on the learnings from Sections~\ref{sec:XNOR-Architectures} and \ref{sec:TwoInternalBlocks}, as well as on the fact that a higher number of internal layers typically increases the accuracy, we propose several architectures (Section~\ref{sec:SeveralInternalBlocks}). Notable are those with accuracy greater than $90\%$ for GTSRB and an average greater than $80\%$ considering also the Belgian and Chinese datasets, and for which the number of parameters varies from 100k to 2M.
\subsection{\textsc{XNOR} Architectures}\label{sec:XNOR-Architectures}
\begin{figure}[h]
  \centering
  \begin{subfigure}[h]{0.4\textwidth}
  \centering
    \includegraphics[width=\textwidth]{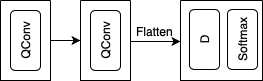}
    \caption{XNOR(QConv) architecture}
    \label{fig:XNOR(QConv)-arch}
  \end{subfigure}
  \hfill
  \begin{subfigure}[h]{0.5\textwidth}
    \includegraphics[width=\textwidth]{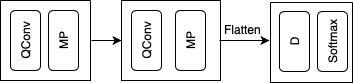}
    \caption{XNOR(QConv,MP) architecture}
    \label{fig:XNOR(QConv,MP)-arch}
  \end{subfigure}
  \caption{XNOR architectures}
  \label{fig:XNOR-architectures}
\end{figure}

We consider the two \textsc{XNOR} architectures from Figure~\ref{fig:XNOR-architectures}. Each is composed of two internal blocks and an output dense (fully connected) layer. Note that, these architectures have only binary parameters. For the GTSRB, the results are in Table~\ref{tab:XNOR(QConv)andXNOR(QConv,MP)}. One could observe that a simple \textsc{XNOR} architecture gives accuracy of at least $70\%$ as long as MP layers are not present but the number of parameters and the model size are high. We can conclude that QConv synthesizes the features well. However, MP layers reduce the accuracy tremendously.
\begin{table}[h]
\caption{\textsc{XNOR(QConv)} and \textsc{XNOR(QConv, MP)} architectures. Image size: 30px $\times$ 30px. Dataset for train and test: GTSRB.}
\label{tab:XNOR(QConv)andXNOR(QConv,MP)}
\centering
\scriptsize
\begin{tabular}{|lcccc|}
\hline
\multicolumn{1}{|l|}{\multirow{2}{*}{\textbf{Model description}}}                                            & \multicolumn{1}{c|}{\multirow{2}{*}{\textbf{Acc}}} & \multicolumn{1}{c|}{\multirow{2}{*}{\textbf{\begin{tabular}[c]{@{}c@{}}\#Binary \\ Params\end{tabular}}}} & \multicolumn{2}{c|}{\textbf{Model Size (in KiB)}}        \\ \cline{4-5} 
\multicolumn{1}{|l|}{}                                                                                       & \multicolumn{1}{c|}{}                              & \multicolumn{1}{c|}{}                                                                                     & \multicolumn{1}{c|}{\textbf{Binary}} & \textbf{Float-32} \\ \hline
\multicolumn{1}{|l|}{\begin{tabular}[c]{@{}l@{}}QConv(32, 3$\times$3),\\ QConv(64, 2$\times$2),\\ D(43)\end{tabular}}            & \multicolumn{1}{c|}{77,91}                         & \multicolumn{1}{c|}{2015264}                                                                              & \multicolumn{1}{c|}{246,5}           & 7874,56           \\ \hline
\multicolumn{1}{|l|}{\begin{tabular}[c]{@{}l@{}}QConv(32, 3$\times$3), MP(2$\times$2),\\ QConv(64, 2$\times$2), MP(2$\times$2),\\ D(43)\end{tabular}}    & \multicolumn{1}{c|}{5,46}                          & \multicolumn{1}{c|}{108128}                                                                               & \multicolumn{1}{c|}{13,2}            & 422,38            \\ \hline
\multicolumn{1}{|l|}{\begin{tabular}[c]{@{}l@{}}QConv(64, 3$\times$3),\\ QConv(128, 2$\times$2), \\ D(43)\end{tabular}}              & \multicolumn{1}{c|}{70,05}                         & \multicolumn{1}{c|}{4046912}                                                                              & \multicolumn{1}{c|}{495,01}          & 15810,56          \\ \hline
\multicolumn{1}{|l|}{\begin{tabular}[c]{@{}l@{}}QConv(64, 3$\times$3), MP(2$\times$2),\\ QConv(128, 2$\times$2), MP(2$\times$2)\\ D(43)\end{tabular}}    & \multicolumn{1}{c|}{10,98}                         & \multicolumn{1}{c|}{232640}                                                                               & \multicolumn{1}{c|}{28,4}            & 908,75            \\ \hline
\multicolumn{1}{|l|}{\begin{tabular}[c]{@{}l@{}}QConv(16, 3$\times$3), \\ QConv(32, 2$\times$2), \\ D(43)\end{tabular}}          & \multicolumn{1}{c|}{81,54}                         & \multicolumn{1}{c|}{1005584}                                                                              & \multicolumn{1}{c|}{122,75}          & 3932,16           \\ \hline
\multicolumn{1}{|l|}{\begin{tabular}[c]{@{}l@{}}QConv(16, 3$\times$3), MP(2$\times$2),\\ QConv(32, 2$\times$2), MP(2$\times$2),\\ D(43)\end{tabular}}    & \multicolumn{1}{c|}{1,42}                          & \multicolumn{1}{c|}{52016}                                                                                & \multicolumn{1}{c|}{6,35}            & 203,19            \\ \hline
\end{tabular}
\end{table}
\subsection{Binarized Neural Architectures}\label{sec:EnhancedXNORArchitectures}
\subsubsection{Two internal blocks}\label{sec:TwoInternalBlocks}
\begin{figure}[h]
  \centering
  \begin{subfigure}[h]{0.4\textwidth}
    \includegraphics[width=\textwidth]{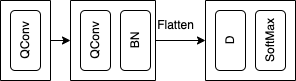}
    \caption{\textsc{XNOR(QConv)} modified}
    \label{fig:modified-XNOR(QConv)-arch}
  \end{subfigure}
  \hfill
  \begin{subfigure}[h]{0.5\textwidth}
    \includegraphics[width=\textwidth]{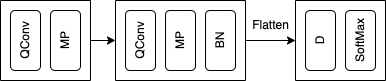}
    \caption{\textsc{XNOR(QConv, MP)} enhanced}
    \label{fig:enhanced-XNOR(QConv)-arch}
  \end{subfigure}
  \caption{BNNs architectures which are not XNOR}
  \label{fig:BNNs-not-XNOR}
\end{figure}

As of Table~\ref{tab:XNOR(QConv)andXNOR(QConv,MP)}, the number of parameters for an architecture with MP layers is at least 15 times less than in a one without, while the size of the binarized models is approx. 30 times less than the 32 bits equivalent. Hence, to benefit from these two sweet spots, we propose a new architecture (see Figure~\ref{fig:enhanced-XNOR(QConv)-arch}) which adds a BN layer in the second block of the \textsc{XNOR} architecture from Figure~\ref{fig:XNOR(QConv,MP)-arch}. The increase in accuracy is considerable (see Table~\ref{tab:Enhanced_XNOR(QConv,MP)_withBN})\footnote{A BN layer following MP is also obtained by composing two blocks of XNOR-Net proposed by ~\cite{rastegari2016xnor}.}. However, a BN layer following a binarized convolution (see Figure~\ref{fig:modified-XNOR(QConv)-arch}) typically leads to a decrease in accuracy (see Table~\ref{tab:Modified_XNOR(QConv)_withBN}). The BN layer introduces few real parameters in the model as well as a slight increase in the model size. This is because only one BN layer was added. Note that the architectures from Figure~\ref{fig:BNNs-not-XNOR} are not XNOR architectures.
\begin{table}[h]
\caption{\textsc{XNOR(QConv, MP)} enhanced. Image size: 30px $\times $30px. Dataset for train and test: GTSRB.}
\label{tab:Enhanced_XNOR(QConv,MP)_withBN}
\centering
\scriptsize
\begin{tabular}{|lcccccc|}
\hline
\multicolumn{1}{|c|}{}                                                                                         & \multicolumn{1}{c|}{}                               & \multicolumn{3}{c|}{\textbf{\#Params}}                                                                                                                     & \multicolumn{2}{c|}{\textbf{Model Size (in KiB)}}        \\ \cline{3-7} 
\multicolumn{1}{|c|}{\multirow{-2}{*}{\textbf{Model description}}}                                             & \multicolumn{1}{c|}{\multirow{-2}{*}{\textbf{Acc}}} & \multicolumn{1}{c|}{\textbf{Binary}}               & \multicolumn{1}{c|}{\textbf{Real}}              & \multicolumn{1}{c|}{\textbf{Total}}                 & \multicolumn{1}{c|}{\textbf{Binary}} & \textbf{Float-32} \\ \hline
\multicolumn{1}{|l|}{\begin{tabular}[c]{@{}l@{}}QConv(32, 3$\times$3), MP(2$\times$2),\\ QConv(64, 2$\times$2), MP(2$\times$2), BN, \\ D(43)\end{tabular}}  & \multicolumn{1}{c|}{50,87}                          & \multicolumn{1}{c|}{{\color[HTML]{212121} 108128}} & \multicolumn{1}{c|}{{\color[HTML]{212121} 128}} & \multicolumn{1}{c|}{{\color[HTML]{212121} 108256}}  & \multicolumn{1}{c|}{13,7}            & 422,88            \\ \hline
\multicolumn{1}{|l|}{\begin{tabular}[c]{@{}l@{}}QConv(64, 3$\times$3), MP(2$\times$2),\\ QConv(128, 2$\times$2), MP(2$\times$2), BN, \\ D(43)\end{tabular}} & \multicolumn{1}{c|}{36,96}                          & \multicolumn{1}{c|}{232640}                        & \multicolumn{1}{c|}{256}                        & \multicolumn{1}{c|}{{\color[HTML]{212121} 232896}}  & \multicolumn{1}{c|}{29,4}            & 909,75            \\ \hline
\multicolumn{1}{|l|}{\begin{tabular}[c]{@{}l@{}}QConv(16, 3$\times$3), MP(2$\times$2),\\ QConv(32, 2$\times$2), MP(2$\times$2), BN, \\ D(43)\end{tabular}}  & \multicolumn{1}{c|}{39,55}                          & \multicolumn{1}{c|}{52016}                         & \multicolumn{1}{c|}{64}                         & \multicolumn{1}{c|}{{\color[HTML]{212121} 52080}}   & \multicolumn{1}{c|}{6,6}             & 203,44            \\ \hline
\end{tabular}
\end{table}
\begin{table}[h]
\caption{\textsc{XNOR(QConv)} modified. Image size: 30px $\times$ 30px. Dataset for train and test: GTSRB.}
\label{tab:Modified_XNOR(QConv)_withBN}
\centering
\scriptsize
\begin{tabular}{|lcccccc|}
\hline
\multicolumn{1}{|c|}{\multirow{2}{*}{\textbf{Model description}}}                                      & \multicolumn{1}{c|}{\multirow{2}{*}{\textbf{Acc}}} & \multicolumn{3}{c|}{\textbf{\#Params}}                                                                          & \multicolumn{2}{c|}{\textbf{Model Size (in KiB)}}        \\ \cline{3-7} 
\multicolumn{1}{|c|}{}                                                                                 & \multicolumn{1}{c|}{}                              & \multicolumn{1}{c|}{\textbf{Binary}} & \multicolumn{1}{c|}{\textbf{Real}} & \multicolumn{1}{c|}{\textbf{Total}} & \multicolumn{1}{c|}{\textbf{Binary}} & \textbf{Float-32} \\ \hline
\multicolumn{1}{|l|}{\begin{tabular}[c]{@{}l@{}}QConv(32, 3$\times$3),\\ QConv(64, 2$\times$2), BN, \\ D(43)\end{tabular}}  & \multicolumn{1}{c|}{82,01}                         & \multicolumn{1}{c|}{2015264}         & \multicolumn{1}{c|}{128}           & \multicolumn{1}{c|}{2015392}        & \multicolumn{1}{c|}{246,5}           & 7874,56           \\ \hline
\multicolumn{1}{|l|}{\begin{tabular}[c]{@{}l@{}}QConv(64, 3$\times$3),\\ QConv(128, 2$\times$2), BN, \\ D(43)\end{tabular}} & \multicolumn{1}{c|}{69,12}                         & \multicolumn{1}{c|}{4046912}         & \multicolumn{1}{c|}{256}           & \multicolumn{1}{c|}{4047168}        & \multicolumn{1}{c|}{495,01}          & 15810,56          \\ \hline
\multicolumn{1}{|l|}{\begin{tabular}[c]{@{}l@{}}QConv(16, 3$\times$3),\\ QConv(32, 2$\times$2), BN, \\ D(43)\end{tabular}}  & \multicolumn{1}{c|}{73,11}                         & \multicolumn{1}{c|}{1005584}         & \multicolumn{1}{c|}{64}            & \multicolumn{1}{c|}{1005648}        & \multicolumn{1}{c|}{123}             & 3932,16           \\ \hline
\end{tabular}
\end{table}
\subsubsection{Several Internal Blocks}\label{sec:SeveralInternalBlocks}
\begin{figure}
  \centering
  \begin{subfigure}[h]{0.47\textwidth}
  \centering
    \includegraphics[width=\textwidth]{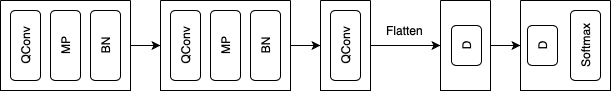}
    \caption{4-blocks Binarized Neural Architecture}
    \label{fig:architecture_v3.drawio.png}
  \end{subfigure}
  \hfill
  \begin{subfigure}[h]{0.47\textwidth}
    \includegraphics[width=\textwidth]{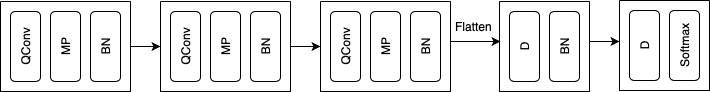}
    \caption{Accuracy-efficient Binarized Neural Architectures}
    \label{fig:architecture_v1.drawio.png}
  \end{subfigure}
  \caption{Binarized Neural Architectures}
  \label{fig:BNA}
\end{figure}

Based on the results obtained in Sections~\ref{sec:XNOR-Architectures} and~\ref{sec:TwoInternalBlocks}, firstly, we trained an architecture where each internal block contains a BN layer only after the MP (see Figure~\ref{fig:architecture_v3.drawio.png}). This is based on the results from Tables~\ref{tab:Enhanced_XNOR(QConv,MP)_withBN} (the BN layer is crucial after MP for accuracy) and \ref{tab:Modified_XNOR(QConv)_withBN} (BN layer after QConv degrades the accuracy). There is an additional internal dense layer for which the number of neurons varies in the set $\{64, 128, 256, 512, 1028\}$. The results are in Table~\ref{tab:QConv_32_5_MP_2_BN_QConv_64_5_MP_2_BN_QConv_64_3_(Dense_???)*_Dense_43}. One could observe that the conclusions drawn from the $2$ blocks architecture do not persist. Hence, motivated also by \cite{hubara2016binarized} we propose the architecture from Figure~\ref{fig:architecture_v1.drawio.png}. 
\begin{table}[h]
\caption{Results for the architecture from the column Model Description. Image size: 30px $\times $30px. Dataset for train and test: GTSRB.}
\label{tab:QConv_32_5_MP_2_BN_QConv_64_5_MP_2_BN_QConv_64_3_(Dense_???)*_Dense_43}
\centering
\scriptsize
\begin{tabular}{|c|c|c|c|ccc|cc|}
\hline
\multirow{2}{*}{\textbf{Model Description}}                                                                                                          & \multirow{2}{*}{\textbf{\#Neur}} & \multirow{2}{*}{\textbf{\#Ep}} & \multirow{2}{*}{\textbf{Acc}} & \multicolumn{3}{c|}{\textbf{\#Params}}                                                                     & \multicolumn{2}{c|}{\textbf{Model size (in KiB)}}                      \\ \cline{5-9} 
                                                                                                                                                     &                                      &                                &                               & \multicolumn{1}{c|}{\textbf{Binary}}         & \multicolumn{1}{c|}{\textbf{Real}}        & \textbf{Total}          & \multicolumn{1}{c|}{\textbf{Binary}}        & \textbf{Float-32}        \\ \hline
\multirow{12}{*}{\begin{tabular}[c]{@{}l@{}}QConv(32, 5x5), MP(2x2), BN,\\ QConv(64, 5x5), MP(2x2), BN,\\ QConv(64, 3x3),\\ D(\#Neur), \\ D(43)\end{tabular}} & \multirow{2}{*}{0}                   & 30                             & 41,17                         & \multicolumn{1}{c|}{\multirow{2}{*}{101472}} & \multicolumn{1}{c|}{\multirow{2}{*}{192}} & \multirow{2}{*}{101664} & \multicolumn{1}{c|}{\multirow{2}{*}{13,14}} & \multirow{2}{*}{397,12}  \\ \cline{3-4}
                                                                                                                                                     &                                      & 100                            & 52,17                         & \multicolumn{1}{c|}{}                        & \multicolumn{1}{c|}{}                     &                         & \multicolumn{1}{c|}{}                       &                          \\ \cline{2-9} 
                                                                                                                                                     & \multirow{2}{*}{64}                  & 30                             & 4,98                          & \multicolumn{1}{c|}{\multirow{2}{*}{109600}} & \multicolumn{1}{c|}{\multirow{2}{*}{192}} & \multirow{2}{*}{109792} & \multicolumn{1}{c|}{\multirow{2}{*}{14,13}} & \multirow{2}{*}{428,88}  \\ \cline{3-4}
                                                                                                                                                     &                                      & 100                            & 5,70                          & \multicolumn{1}{c|}{}                        & \multicolumn{1}{c|}{}                     &                         & \multicolumn{1}{c|}{}                       &                          \\ \cline{2-9} 
                                                                                                                                                     & \multirow{2}{*}{128}                 & 30                             & 7,03                          & \multicolumn{1}{c|}{\multirow{2}{*}{128736}} & \multicolumn{1}{c|}{\multirow{2}{*}{192}} & \multirow{2}{*}{128928} & \multicolumn{1}{c|}{\multirow{2}{*}{16,46}} & \multirow{2}{*}{503,62}  \\ \cline{3-4}
                                                                                                                                                     &                                      & 100                            & 5,70                          & \multicolumn{1}{c|}{}                        & \multicolumn{1}{c|}{}                     &                         & \multicolumn{1}{c|}{}                       &                          \\ \cline{2-9} 
                                                                                                                                                     & \multirow{2}{*}{256}                 & 30                             & 12,43                         & \multicolumn{1}{c|}{\multirow{2}{*}{167008}} & \multicolumn{1}{c|}{\multirow{2}{*}{192}} & \multirow{2}{*}{167200} & \multicolumn{1}{c|}{\multirow{2}{*}{21,14}} & \multirow{2}{*}{653,12}  \\ \cline{3-4}
                                                                                                                                                     &                                      & 100                            & 8,48                          & \multicolumn{1}{c|}{}                        & \multicolumn{1}{c|}{}                     &                         & \multicolumn{1}{c|}{}                       &                          \\ \cline{2-9} 
                                                                                                                                                     & \multirow{2}{*}{512}                 & 30                             & 19,82                         & \multicolumn{1}{c|}{\multirow{2}{*}{243552}} & \multicolumn{1}{c|}{\multirow{2}{*}{192}} & \multirow{2}{*}{243744} & \multicolumn{1}{c|}{\multirow{2}{*}{30,48}} & \multirow{2}{*}{952,12}  \\ \cline{3-4}
                                                                                                                                                     &                                      & 100                            & 32,13                         & \multicolumn{1}{c|}{}                        & \multicolumn{1}{c|}{}                     &                         & \multicolumn{1}{c|}{}                       &                          \\ \cline{2-9} 
                                                                                                                                                     & \multirow{2}{*}{1024}                & 30                             & 46,05                         & \multicolumn{1}{c|}{\multirow{2}{*}{396640}} & \multicolumn{1}{c|}{\multirow{2}{*}{192}} & \multirow{2}{*}{396832} & \multicolumn{1}{c|}{\multirow{2}{*}{49,17}} & \multirow{2}{*}{1546,24} \\ \cline{3-4}
                                                                                                                                                     &                                      & 100                            & 50,91                         & \multicolumn{1}{c|}{}                        & \multicolumn{1}{c|}{}                     &                         & \multicolumn{1}{c|}{}                       &                          \\ \hline
\end{tabular}
\end{table}
\section{Experimental results and discussion}\label{sec:expResults}
\begin{figure}[h]
  \centering
    \includegraphics[width=0.7\textwidth]{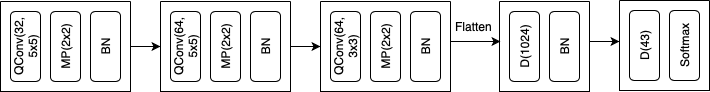}
    \caption{Accuracy Efficient Architecture for GTSRB and Belgium dataset}
    \label{fig:Acc-Efficient-Arch-GTSRB-Belgium}
\end{figure}

\begin{figure}[h]
  \centering
    \includegraphics[width=0.7\textwidth]{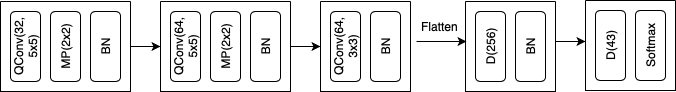}
    \caption{Accuracy Efficient Architecture for Chinese dataset}
    \label{fig:Acc-Efficient-Arch-Chinese}
\end{figure}

The best accuracy for GTSRB and Belgium datasets is $96,45$ and $88,17$, respectively, and was obtained for the architecture from Figure~\ref{fig:Acc-Efficient-Arch-GTSRB-Belgium}, with input size 64$\times$64 (see Table~\ref{tab:QConv_32_5_MP_2_BN_QConv_64_5_MP_2_BN_QConv_64_3_MP_2_BN_(Dense_???)*_Dense_43}). The number of parameters is almost $2$M and the model size $225,67$~KiB (for the binary model) and $6932,48$ KiB (for the Float-32 equivalent). 
There is no surprise the same architecture gave the best results for GTSRB and Belgium since they belong to the European area. 
\begin{table}[h]
\centering
\scriptsize
\caption{Results for the architecture from Figure~\ref{fig:Acc-Efficient-Arch-GTSRB-Belgium}. Dataset for train: GTSRB.}
\label{tab:QConv_32_5_MP_2_BN_QConv_64_5_MP_2_BN_QConv_64_3_MP_2_BN_(Dense_???)*_Dense_43}
\begin{tabular}{|l|c|ccc|ccc|cc|}
\hline
\multirow{2}{*}{\textbf{Input size}} &
  \multicolumn{1}{c|}{\multirow{2}{*}{\textbf{\#Neur}}} &
  \multicolumn{3}{c|}{\textbf{Accuracy}} &
  \multicolumn{3}{c|}{\textbf{\#Params}} &
  \multicolumn{2}{c|}{\textbf{Model Size (in KiB)}} \\ \cline{3-10} 
 &
  \multicolumn{1}{c|}{} &
  \multicolumn{1}{c|}{\textbf{German}} &
  \multicolumn{1}{c|}{\textbf{China}} &
  \multicolumn{1}{c|}{\textbf{Belgium}} &
  \multicolumn{1}{c|}{\textbf{Binary}} &
  \multicolumn{1}{c|}{\textbf{Real}} &
  \multicolumn{1}{c|}{\textbf{Total}} &
  \multicolumn{1}{c|}{\textbf{Binary}} &
  \multicolumn{1}{c|}{\textbf{Float-32}} \\ \hline
\multirow{6}{*}{64px $\times$ 64px} &
  0 &
  \multicolumn{1}{c|}{93,83} &
  \multicolumn{1}{c|}{77,86} &
  79,75 &
  \multicolumn{1}{c|}{159264} &
  \multicolumn{1}{c|}{320} &
  159584 &
  \multicolumn{1}{c|}{20,69} &
  623,38 \\ \cline{2-10} 
 &
  64 &
  \multicolumn{1}{c|}{94,43} &
  \multicolumn{1}{c|}{75,09} &
  82,39 &
  \multicolumn{1}{c|}{195616} &
  \multicolumn{1}{c|}{448} &
  196064 &
  \multicolumn{1}{c|}{25,63} &
  765,88 \\ \cline{2-10} 
 &
  128 &
  \multicolumn{1}{c|}{95,42} &
  \multicolumn{1}{c|}{74,71} &
  83,44 &
  \multicolumn{1}{c|}{300768} &
  \multicolumn{1}{c|}{576} &
  301344 &
  \multicolumn{1}{c|}{38,96} &
  1177,60 \\ \cline{2-10} 
 &
  256 &
  \multicolumn{1}{c|}{94,75} &
  \multicolumn{1}{c|}{80,37} &
  81,40 &
  \multicolumn{1}{c|}{511072} &
  \multicolumn{1}{c|}{832} &
  511904 &
  \multicolumn{1}{c|}{65,64} &
  1996,80 \\ \cline{2-10} 
 &
  512 &
  \multicolumn{1}{c|}{95,65} &
  \multicolumn{1}{c|}{78,49} &
  85,64 &
  \multicolumn{1}{c|}{931680} &
  \multicolumn{1}{c|}{1344} &
  933024 &
  \multicolumn{1}{c|}{118,98} &
  3645,44 \\ \cline{2-10} 
 &
  1024 &
  \multicolumn{1}{c|}{\textbf{96,45}} &
  \multicolumn{1}{c|}{\textbf{81,50}} &
  \textbf{88,17} &
  \multicolumn{1}{c|}{1772896} &
  \multicolumn{1}{c|}{2368} &
  1775264 &
  \multicolumn{1}{c|}{225,67} &
  6932,48 \\ \hline
\end{tabular}
\end{table}
The best accuracy for Chinese dataset ($83,9\%$) is obtained by another architecture, namely from Figure~\ref{fig:Acc-Efficient-Arch-Chinese}, with input size 48$\times$48 (see Table~\ref{tab:QConv_32_5_MP_2_BN_QConv_64_5_MP_2_BN_QConv_64_3_BN_(Dense_???)*_Dense_43}). This architecture is more efficient from the point of view of computationally limited devices and formal verification having $900$k parameters and $113,64$ KiB (for the binary model) and $3532,8$ KiB (for the Float-32 equivalent). Also, the second architecture gave the best average accuracy and the decrease in accuracy for GTSRB and Belgium is small, namely $1,17\%$ and $0,39\%$, respectively.
\begin{table}[h]
\centering
\scriptsize
\caption{Results for the architecture from Figure~\ref{fig:Acc-Efficient-Arch-Chinese}. Dataset for train: GTSRB.}
\label{tab:QConv_32_5_MP_2_BN_QConv_64_5_MP_2_BN_QConv_64_3_BN_(Dense_???)*_Dense_43}
\begin{tabular}{|l|c|ccc|ccc|cc|}
\hline
\multirow{2}{*}{\textbf{Input size}} &
  \multicolumn{1}{c|}{\multirow{2}{*}{\textbf{\#Neur}}} &
  \multicolumn{3}{c|}{\textbf{Accuracy}} &
  \multicolumn{3}{c|}{\textbf{\#Params}} &
  \multicolumn{2}{c|}{\textbf{Model Size (in KiB)}} \\ \cline{3-10} 
 &
  \multicolumn{1}{c|}{} &
  \multicolumn{1}{c|}{\textbf{German}} &
  \multicolumn{1}{c|}{\textbf{China}} &
  \multicolumn{1}{c|}{\textbf{Belgium}} &
  \multicolumn{1}{c|}{\textbf{Binary}} &
  \multicolumn{1}{c|}{\textbf{Real}} &
  \multicolumn{1}{c|}{\textbf{Total}} &
  \multicolumn{1}{c|}{\textbf{Binary}} &
  \multicolumn{1}{c|}{\textbf{Float-32}} \\ \hline
\multirow{6}{*}{48px $\times$ 48px} &
  0 &
  \multicolumn{1}{c|}{94,67} &
  \multicolumn{1}{c|}{82,13} &
  83,16 &
  \multicolumn{1}{c|}{225312} &
  \multicolumn{1}{c|}{320} &
  225632 &
  \multicolumn{1}{c|}{28,75} &
  881,38 \\ \cline{2-10} 
 &
  64 &
  \multicolumn{1}{c|}{94,56} &
  \multicolumn{1}{c|}{82,38} &
  85,75 &
  \multicolumn{1}{c|}{293920} &
  \multicolumn{1}{c|}{448} &
  294368 &
  \multicolumn{1}{c|}{37,63} &
  1146,88 \\ \cline{2-10} 
 &
  128 &
  \multicolumn{1}{c|}{95,02} &
  \multicolumn{1}{c|}{81,50} &
  87,45 &
  \multicolumn{1}{c|}{497376} &
  \multicolumn{1}{c|}{576} &
  497952 &
  \multicolumn{1}{c|}{62,96} &
  1945,60 \\ \cline{2-10} 
 &
  256 &
  \multicolumn{1}{c|}{\textbf{95,28}} &
  \multicolumn{1}{c|}{\textbf{83,90}} &
  \textbf{87,78} &
  \multicolumn{1}{c|}{904288} &
  \multicolumn{1}{c|}{832} &
  905120 &
  \multicolumn{1}{c|}{113,64} &
  3532,80 \\ \cline{2-10} 
 &
  512 &
  \multicolumn{1}{c|}{95,90} &
  \multicolumn{1}{c|}{76,22} &
  87,34 &
  \multicolumn{1}{c|}{1718112} &
  \multicolumn{1}{c|}{1344} &
  1719456 &
  \multicolumn{1}{c|}{214,98} &
  6717,44 \\ \cline{2-10} 
 &
  1024 &
  \multicolumn{1}{c|}{95,37} &
  \multicolumn{1}{c|}{81,76} &
  86,74 &
  \multicolumn{1}{c|}{3345760} &
  \multicolumn{1}{c|}{2368} &
  3348128 &
  \multicolumn{1}{c|}{417,67} &
  13076,48 \\ \hline
\end{tabular}
\end{table}

If we investigate both architectures based on confusion matrix results, for GTSRB we observe that the model failed to predict, for example, the \emph{End of speed limit 80} and \emph{Bicycle Crossing}. The first was confused the most with \emph{Speed limit (80km/h)}, the second with \emph{Children crossing}. One reason for the first confusion could be that \emph{End of speed limit (80 km/h)} might be considered the occluded version of \emph{Speed limit (80km/h)}. 

\begin{figure}[h!]
  \centering
  \begin{subfigure}[h]{0.47\textwidth}
  \centering
    \includegraphics[width=\textwidth]{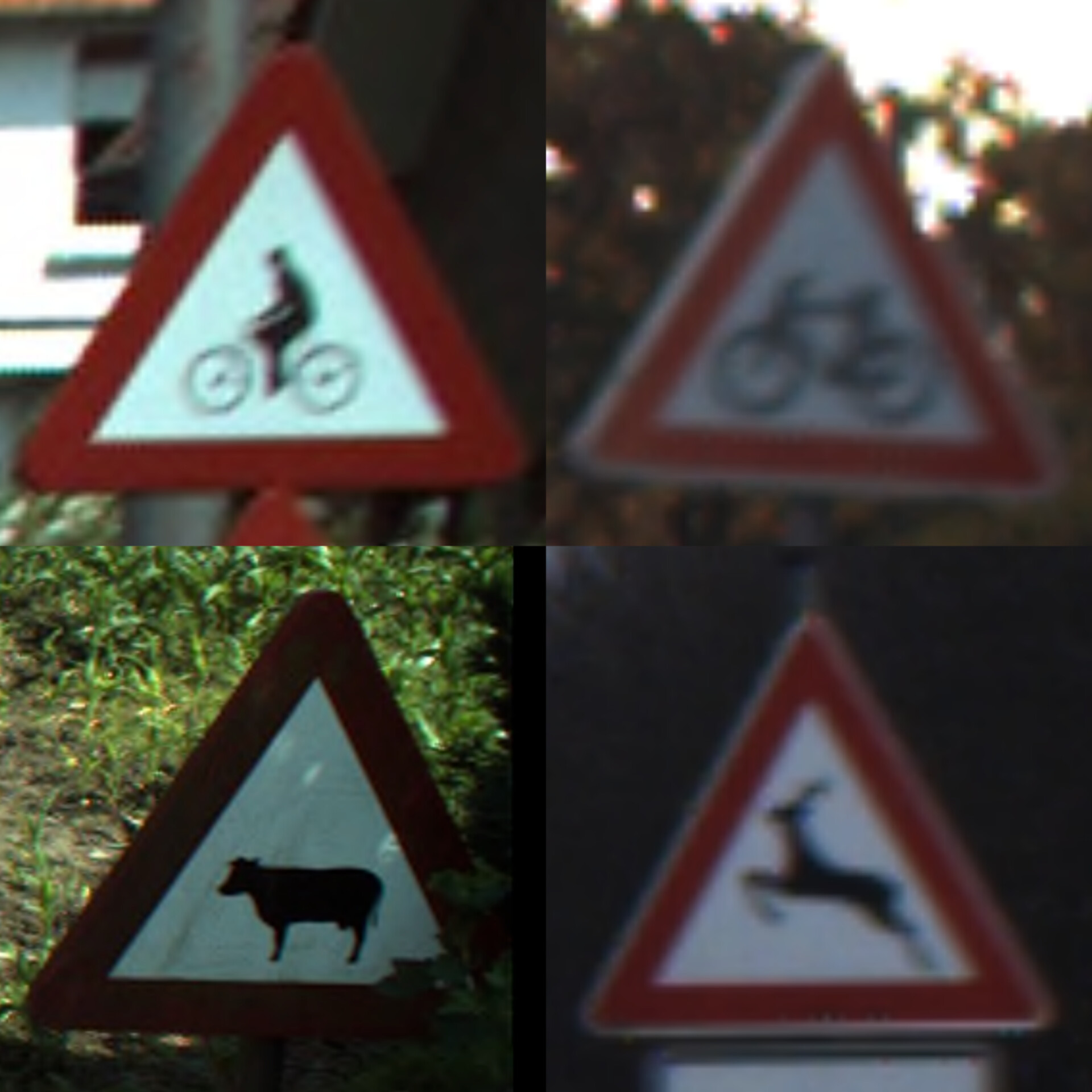}
    \caption{Difference between Belgium (left) and GRSRB (right) dataset}
    \label{fig:belgium_german}
  \end{subfigure}
  \hfill
  \begin{subfigure}[h]{0.47\textwidth}
    \includegraphics[width=\textwidth]{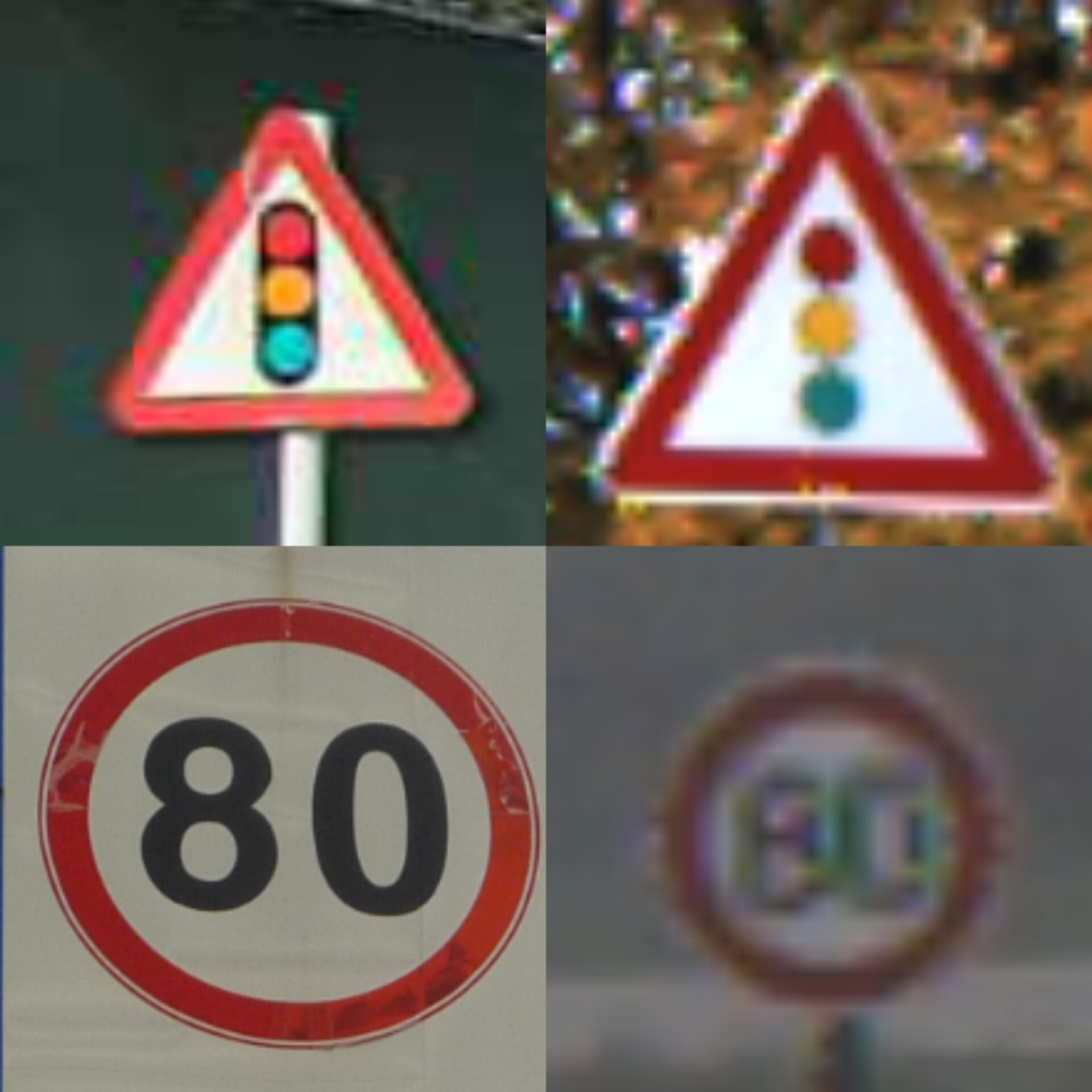}
    \caption{Difference between Chinese (left) and GRSRB (right) dataset}
    \label{fig:chinese_german}
  \end{subfigure}
\caption{Differences between traffic sign in the datasets}
\label{fig:Differences-traffic-sign-datasets}
\end{figure}

For Belgium test set, the worst results were obtained, for example, for \emph{Bicycle crossing} and \emph{Wild animals crossing} because the images differ a lot from the images on GTSRB training set (see Figure \ref{fig:belgium_german}). Another bad prediction is for \emph{Double Curve} which was equally confused with \emph{Slippery road} and \emph{Children crossing}.

In the Chinese test set, the \emph{Traffic signals} failed to be predicted at all by the model proposed by us and was assimilated with the \emph{General Caution} class from the GTSRB, however \emph{General Caution} is not a class in the Chinese test set (see Figure \ref{fig:chinese_german}, top). Another bad prediction is for \emph{Speed limit (80km/h)} which was equally confused with \emph{Speed limit (30km/h), Speed limit (50km/h)} and \emph{Speed limit (60km/h)} but not with \emph{Speed limit (70km/h)}. One reason could be the quality of the training images compared to the test ones  (see Figure \ref{fig:chinese_german}, bottom).

In conclusion, there are few cases when the prediction failures can be explained, however the need for formal verification guarantees of the results is urgent which we will be performed as future work.

\begin{thebibliography}{10}
\providecommand{\url}[1]{\texttt{#1}}
\providecommand{\urlprefix}{URL }
\providecommand{\doi}[1]{https://doi.org/#1}

\bibitem{BelgianTrafficSignDatabase}
{Belgian Traffic Sign Database}.
  \url{https://www.kaggle.com/datasets/shazaelmorsh/trafficsigns}, accessed:
  March 25th, 2023

\bibitem{BenchmarksVNN-COMP-22}
{Benchmarks of the 3rd International Verification of Neural Networks
  Competition (VNN-COMP'22)}.
  \url{https://github.com/ChristopherBrix/vnncomp2022_benchmarks}, accessed:
  February 22nd, 2023

\bibitem{ChineseTrafficSignDatabase}
{Chinese Traffic Sign Database}.
  \url{https://www.kaggle.com/datasets/dmitryyemelyanov/chinese-traffic-signs},
  accessed: March 25th, 2023

\bibitem{GTSRB}
{German Traffic Sign Recognition Benchmark}.
  \url{https://www.kaggle.com/datasets/meowmeowmeowmeowmeow/gtsrb-german-traffic-sign?datasetId=82373&language=Python},
  accessed: March 25th, 2023

\bibitem{amir2021smt}
Amir, G., Wu, H., Barrett, C., Katz, G.: {An SMT-based Approach for Verifying
  Binarized Neural Networks}. In: Tools and Algorithms for the Construction and
  Analysis of Systems: 27th International Conference, TACAS 2021, Held as Part
  of the European Joint Conferences on Theory and Practice of Software, ETAPS
  2021, Luxembourg City, Luxembourg, March 27--April 1, 2021, Proceedings, Part
  II 27. pp. 203--222. Springer (2021)

\bibitem{chen2019investigating}
Chen, E.H., R{\"o}thig, P., Zeisler, J., Burschka, D.: {Investigating Low Level
  Features in CNN for Traffic Sign Detection and Recognition}. In: 2019 IEEE
  Intelligent Transportation Systems Conference (ITSC). pp. 325--332. IEEE
  (2019)

\bibitem{chen2021investigating}
Chen, E.H., Vemparala, M.R., Fasfous, N., Frickenstein, A., Mzid, A., Nagaraja,
  N.S., Zeisler, J., Stechele, W.: {Investigating Binary Neural Networks for
  Traffic Sign Detection and Recognition}. In: 2021 IEEE Intelligent Vehicles
  Symposium (IV). pp. 1400--1405. IEEE (2021)

\bibitem{ciregan2012multi}
Ciregan, D., Meier, U., Schmidhuber, J.: {Multi-Column Deep Neural Networks for
  Image Classification}. In: 2012 IEEE conference on computer vision and
  pattern recognition. pp. 3642--3649. IEEE (2012)

\bibitem{courbariaux2015binaryconnect}
Courbariaux, M., Bengio, Y., David, J.P.: {BinaryConnect: Training Deep Neural
  Networks with Binary Weights during Propagations}. Advances in neural
  information processing systems  \textbf{28} (2015)

\bibitem{deng2009imagenet}
Deng, J., Dong, W., Socher, R., Li, L.J., Li, K., Fei-Fei, L.: {ImageNet: A
  Large-Scale Hierarchical Image Database}. In: 2009 IEEE conference on
  computer vision and pattern recognition. pp. 248--255. Ieee (2009)

\bibitem{geiger2020larq}
Geiger, L., Team, P.: {Larq: An Open-Source Library for Training Binarized
  Neural Networks}. Journal of Open Source Software  \textbf{5}(45), ~1746
  (2020)

\bibitem{guo2023occrob}
Guo, X., Zhou, Z., Zhang, Y., Katz, G., Zhang, M.: {OccRob: Efficient SMT-Based
  Occlusion Robustness Verification of Deep Neural Networks}. arXiv preprint
  arXiv:2301.11912  (2023)

\bibitem{houben2013detection}
Houben, S., Stallkamp, J., Salmen, J., Schlipsing, M., Igel, C.: {Detection of
  Traffic Signs in Real-World Images: The German Traffic Sign Detection
  Benchmark}. In: The 2013 international joint conference on neural networks
  (IJCNN). pp.~1--8. Ieee (2013)

\bibitem{hubara2016binarized}
Hubara, I., Courbariaux, M., Soudry, D., El-Yaniv, R., Bengio, Y.: {Binarized
  Neural Networks}. Advances in neural information processing systems
  \textbf{29} (2016)

\bibitem{hubara2017quantized}
Hubara, I., Courbariaux, M., Soudry, D., El-Yaniv, R., Bengio, Y.: {Quantized
  Neural Networks: Training Neural Networks with Low Precision Weights and
  Activations}. The Journal of Machine Learning Research  \textbf{18}(1),
  6869--6898 (2017)

\bibitem{ioffe2015batch}
Ioffe, S., Szegedy, C.: {Batch Normalization: Accelerating Deep Network
  Training by Reducing Internal Covariate Shift}. In: International conference
  on machine learning. pp. 448--456. PMLR (2015)

\bibitem{katz2017reluplex}
Katz, G., Barrett, C., Dill, D.L., Julian, K., Kochenderfer, M.J.: {Reluplex:
  An Efficient SMT Solver for Verifying Deep Neural Networks}. In: Computer
  Aided Verification: 29th International Conference, CAV 2017, Heidelberg,
  Germany, July 24-28, 2017, Proceedings, Part I 30. pp. 97--117. Springer
  (2017)

\bibitem{krishnamoorthi2018quantizing}
Krishnamoorthi, R.: {Quantizing Deep Convolutional Networks for Efficient
  Inference: A whitepaper}. arXiv preprint arXiv:1806.08342  (2018)

\bibitem{krizhevsky2009learning}
Krizhevsky, A., Hinton, G., et~al.: {Learning Multiple Layers of Features from
  Tiny Images}  (2009)

\bibitem{lecun1998mnist}
LeCun, Y.: {The MNIST Database of Handwritten Digits}. http://yann. lecun.
  com/exdb/mnist/  (1998)

\bibitem{lin2017focal}
Lin, T.Y., Goyal, P., Girshick, R., He, K., Doll{\'a}r, P.: {Focal Loss for
  Dense Object Detection}. In: Proceedings of the IEEE international conference
  on computer vision. pp. 2980--2988 (2017)

\bibitem{10.1007/978-3-540-78800-3_24}
de~Moura, L., Bj{\o}rner, N.: {Z3: An Efficient {SMT} Solver}. In:
  Ramakrishnan, C.R., Rehof, J. (eds.) Tools and Algorithms for the
  Construction and Analysis of Systems. pp. 337--340. Springer Berlin
  Heidelberg, Berlin, Heidelberg (2008)

\bibitem{narodytska2018formal}
Narodytska, N.: {Formal Analysis of Deep Binarized Neural Networks}. In: IJCAI.
  pp. 5692--5696 (2018)

\bibitem{netzer2011reading}
Netzer, Y., Wang, T., Coates, A., Bissacco, A., Wu, B., Ng, A.Y.: {Reading
  Digits in Natural Images with Unsupervised Feature Learning}  (2011)

\bibitem{rastegari2016xnor}
Rastegari, M., Ordonez, V., Redmon, J., Farhadi, A.: {XNOR-Net: ImageNet
  Classification using Binary Convolutional Neural Networks}. In: European
  conference on computer vision. pp. 525--542. Springer (2016)

\bibitem{ruder2016overview}
Ruder, S.: {An Overview of Gradient Descent Optimization Algorithms}. arXiv
  preprint arXiv:1609.04747  (2016)

\bibitem{sermanet2011traffic}
Sermanet, P., LeCun, Y.: {Traffic Sign Recognition with Multi-Scale
  Convolutional Networks}. In: The 2011 international joint conference on
  neural networks. pp. 2809--2813. IEEE (2011)

\bibitem{simonyan2014very}
Simonyan, K., Zisserman, A.: {Very Deep Convolutional Networks for Large-Scale
  Image Recognition}. arXiv preprint arXiv:1409.1556  (2014)

\bibitem{szegedy2013intriguing}
Szegedy, C., Zaremba, W., Sutskever, I., Bruna, J., Erhan, D., Goodfellow, I.,
  Fergus, R.: {Intriguing Properties of Neural Networks}. arXiv preprint
  arXiv:1312.6199  (2013)

\bibitem{zhang2020lightweight}
Zhang, J., Wang, W., Lu, C., Wang, J., Sangaiah, A.K.: {Lightweight Deep
  Network for Traffic Sign Classification}. Annals of Telecommunications
  \textbf{75},  369--379 (2020)

\end{thebibliography}

\end{document}